\documentclass[letterpaper]{article} 
\usepackage{aaai2027}  
\usepackage[hyphens]{url}  
\usepackage{graphicx} 
\usepackage{natbib}  
\usepackage{caption} 
\usepackage{amsmath}
\usepackage{amssymb}

\usepackage{algorithm}
\usepackage{algorithmic}

\usepackage{newfloat}
\usepackage{listings}
\DeclareCaptionStyle{ruled}{labelfont=normalfont,labelsep=colon,strut=off} 
\floatstyle{ruled}
\newfloat{listing}{tb}{lst}{}
\floatname{listing}{Listing}

\usepackage{booktabs}

\nocopyright
\title{Between Gradient and Natural Gradient: A Continuum of LoRA Initializations}
\author{
    Dianze Liu\equalcontrib,
    Farshid Ghezelbash\equalcontrib
}
\affiliations{
    Georgia Institute of Technology\\
    dliu450@gatech.edu, ghezelbash.far@gmail.com
}

\begin{document}

\maketitle

\begin{abstract}
Low-rank adaptation (LoRA) fine-tunes large pretrained models at a
fraction of the cost of full fine-tuning, but its performance depends
strongly on how the adapters are initialized. Recent schemes
initialize the adapters from the downstream loss gradient: some project
the raw gradient onto its top directions, while others first whiten it
with an estimate of the loss curvature. We show that these seemingly
distinct methods are points on a single continuum: a two-parameter
family of preconditioned gradient initializations, which we call Unified
LoRA (ULoRA), governed by a spectral whitening exponent and an
Adam-like diagonal exponent. Sweeping this family under a full learning-rate search, we
find that no single fixed preconditioning strength dominates: the best
operating point is task-dependent and frequently lies strictly inside
the family, away from the published endpoints. Treated as an
upper bound of this family, a tuned ULoRA configuration matches or
exceeds full fine-tuning on all five GLUE tasks with RoBERTa-base and is
competitive with the strongest baselines on GSM8K with LLaMA~2-7B. Our
deployable, search-free variant, ULoRA-Auto, selects per-layer exponents
from measured spectral statistics, approaches this upper bound at no
additional search cost, and ranks at or near the top among deployable
LoRA methods. Our results show that a principled design space for LoRA
initialization and curvature preconditioning should be treated
as a tunable dimension rather than a fixed design decision.
\end{abstract}



\section{Introduction}
Adapting large pre-trained models to downstream tasks has become standard practice. Full fine-tuning achieves strong performance but updates all parameters, which is expensive and often unnecessary. Low-rank adaptation (LoRA) \cite{hu2021LoRA} addresses this by restricting weight updates to a low-rank form $\Delta W = \frac{1}{\sqrt{r}} BA$, where only two small matrices $A \in \mathbb{R}^{r \times d_\text{in}}$ and $B \in \mathbb{R}^{d_\text{out} \times r}$ are trained. This reduces trainable parameters by orders of magnitude while keeping the pretrained backbone frozen.
 
Despite their efficiency, LoRA methods could underperform full fine-tuning, particularly on small datasets and complex tasks. A key factor is the initialization of $A$ and $B$: standard LoRA uses $B = 0$ and random $A$, which provides no information about the target task. Recent work has shown that choosing the initial subspace from task data substantially narrows the performance gap. LoRA-GA \cite{wang2024LoRAga} and LoRA-One  \cite{zhang2025LoRA1} align the adapters with the top directions of the loss gradient. CG-LoRA \cite{zheng2026cg-LoRA} further whitens the gradient using Kronecker-factored curvature (K-FAC).
 
\paragraph{Gap.} Despite this progress, existing gradient-based methods make a rigid choice about how the gradient is used for initialization. Gradient-only methods (LoRA-GA, LoRA-One) ignore curvature entirely, which overlooks second-order information about the loss landscape. Curvature-whitened methods (CG-LoRA) apply a fixed, theoretically motivated whitening exponent of $-1/2$, which can be too aggressive when curvature estimates are noisy, as is common in practice with small batches and nonlinear models. The question of whether a different level of curvature preconditioning would be better has not been studied.
 
\begin{figure*}[t]
\centering
\includegraphics[width=1\textwidth]{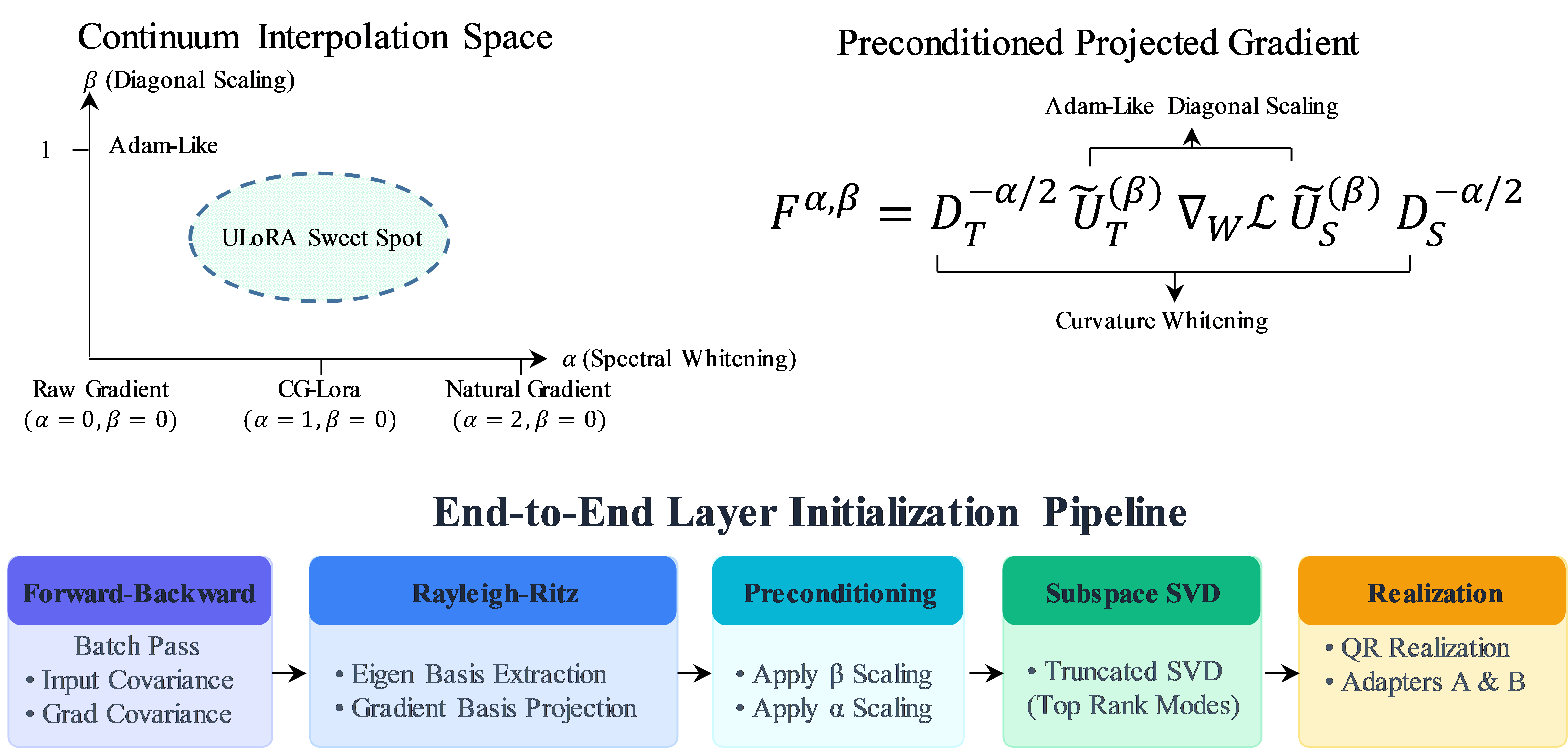}
\caption{Overview of ULoRA. Top left: the continuum interpolation space spanned by spectral whitening ($\alpha$) and diagonal scaling ($\beta$), with existing methods (e.g., CG-LoRA). Top right: the preconditioned projected gradient $F^{\alpha,\beta}$. Bottom: schematics of the entire pipeline.}
\label{fig:overview}
\end{figure*}

\paragraph{Our approach.} We propose Unified LoRA (ULoRA; Figure \ref{fig:overview}), which shows that all existing gradient-based initialization methods are special cases of a common framework parametrized by two scalars:
 
\begin{itemize}
    \item  Spectral whitening exponent ($\alpha$), which controls how strongly K-FAC eigenvalues rescale gradient directions. The value $\alpha = 0$ removes curvature scaling; $\alpha = 1$ recovers CG-LoRA. 
    \item  Diagonal preconditioning exponent ($\beta$), which applies a normalization by the empirical variance of layer inputs and output gradients, analogous to the second-moment scaling in Adam \cite{kingma2014adam}.
\end{itemize}

This two-parameter family reproduces gradient projection ($\alpha = 0, \beta = 0$), Adam-like preconditioning ($\alpha = 0, \beta > 0$), full curvature whitening ($\alpha = 1, \beta = 0$), and mixed strategies that combine both. Following a systematic grid search over $(\alpha, \beta)$ we find that rather than either of the two major families dominating, intermediate configurations frequently achieve superior performance across all evaluated tasks.
 
 
\paragraph{Contributions.} Our contributions are as follows: 
\begin{enumerate}
    \item We propose ULoRA, a two-parameter family of LoRA initialization strategies that unifies existing gradient-based and curvature-based methods under a single framework.
 
    \item We provide a characterization of $\alpha$ and $\beta$ and show how they recover known methods as special cases, establishing a principled design space for LoRA initialization.
    
    \item We demonstrate that the optimal level of curvature preconditioning is neither zero nor the full value, and that intermediate values are consistently more robust across various language benchmarks.
 
\end{enumerate}
Treated as an oracle upper bound of the family, a tuned ULoRA configuration matches or exceeds full fine-tuning on all five GLUE tasks with RoBERTa-base, while the deployable, search-free ULoRA-Auto recovers most of this gain and ranks at or near the top among deployable LoRA methods on GSM8K with LLaMA~2-7B and across GLUE.

\section{Related Works}
 
\paragraph{Variants of vanilla LoRA.}
Standard LoRA \cite{hu2021LoRA} initializes $B = 0$ and $A$ randomly, so fine-tuning starts from zero perturbation of the pretrained model. A first line of follow-up work improves how the adapters are trained rather than where they start: rsLoRA \cite{kalajdzievski2023rank} rescales the update by $\gamma/\sqrt{r}$ to stabilize high ranks, LoRA+ \cite{hayou2024LoRA+} assigns separate learning rates to $A$ and $B$, DoRA \cite{liu2024dora} decomposes weight updates into magnitude and directional components, and AdaLoRA \cite{zhang2023adaLoRA} reallocates the rank budget across layers by parameter importance. None of these methods use task data to choose the initial subspace, which is the axis on which ULoRA operates.

\paragraph{Pretrained-informed initialization.}
A second line of work narrows the gap to full fine-tuning by making the initialization itself data-dependent. Early methods draw the subspace from the model rather than the task: PiSSA \cite{meng2024pissa} initializes from the principal singular components of the pretrained weight $W_0$, LoftQ \cite{li2024loftq} jointly optimizes quantization and low-rank initialization, and EVA \cite{paischer2025eva} uses the SVD of activation vectors to maximize captured activation variance. These subspaces reflect what the pretrained model already represents, but not the direction in which the downstream loss wants to move it.
 
\paragraph{Gradient-based initialization.}
The methods most closely related to ULoRA derive the subspace directly from the downstream loss gradient. LoRA-GA \cite{wang2024LoRAga} aligns the adapters with the first-step full fine-tuning gradient, and LoRA-One  \cite{zhang2025LoRA1} proves that a single full-gradient step yields near-optimal subspace alignment under mild assumptions. CG-LoRA \cite{zheng2026cg-LoRA} goes one step further: rather than using the raw gradient, it whitens the gradient with Kronecker-factored curvature (K-FAC) \cite{martens2015optimizing} before extracting the subspace, motivated by a function-space alignment objective. These methods therefore differ in exactly one design decision, namely how much curvature information reshapes the gradient before the subspace is extracted: none at all (LoRA-GA, LoRA-One) or full inverse-square-root whitening (CG-LoRA). ULoRA makes this decision explicit and continuous, parameterizing the spectrum between unwhitened gradient projection, Adam-like diagonal normalization, and full spectral whitening, and showing that the best operating point lies strictly between the published extremes.
 
\paragraph{Learning rate sensitivity:} LoRA is highly sensitive to learning rate choice, and a well-tuned vanilla LoRA is competitive with more complex methods \cite{lee2026learning}. This motivates our full learning rate sweep across all methods and configurations.

\section{Method}
 
\subsection{Preconditioned Gradients: A Unifying View}
Optimizers differ mostly in how they precondition the gradient
$\nabla_W \mathcal{L}$ before taking a step. Plain gradient descent
uses the raw gradient, $\Delta W \propto \nabla_W \mathcal{L}$. Adam \cite{kingma2014adam}
rescales each coordinate by the inverse root of its second moment,
$\Delta W \propto \mathrm{diag}(v)^{-1/2} \odot \nabla_W \mathcal{L}$. Natural gradient
descent uses the full curvature, $\Delta W \propto F^{-1}\nabla\mathcal{L}$
\cite{amari1998natural}; under the Kronecker factorization
$F \approx T \otimes S$ of K-FAC \cite{martens2015optimizing}, this
becomes $\Delta W \propto T^{-1} \nabla_W \mathcal{L}\, S^{-1}$. Shampoo
\cite{gupta2018shampoo} interpolates between these regimes with
quarter-power factors, $\Delta W \propto T^{-1/4} \nabla_W \mathcal{L}\, S^{-1/4}$.
All of these are instances of a single two-parameter family:
\begin{equation}
\label{eq:family}
P_{\alpha,\beta} \;=\;
T^{-\alpha/2}
\Big[ d_T^{-\beta/2} \odot \nabla_W \mathcal{L} \odot d_S^{-\beta/2} \Big]
S^{-\alpha/2},
\end{equation}
where $\alpha$ controls the strength of Kronecker-factored curvature
whitening, $\beta$ controls Adam-like per-feature diagonal
normalization, and $d_S, d_T$ are the diagonals of $S, T$. Setting
$(\alpha,\beta)=(0,0)$ recovers the raw gradient; $(0,\beta)$
recovers Adam-like diagonal scaling; $(2,0)$ recovers the K-FAC natural
gradient; $(0.5,0)$ matches Shampoo's quarter-power preconditioning;
and $(1,0)$ recovers inverse-square-root whitening.
 
Our key observation is that gradient-based LoRA initialization faces the same design choice. LoRA-GA \cite{wang2024LoRAga} and LoRA-One \cite{zhang2025LoRA1} extract the adapter subspace from the raw gradient, i.e.\ $P_{0,0}$, while CG-LoRA \cite{zheng2026cg-LoRA}
extracts it from the whitened gradient $T^{-1/2} \nabla_W \mathcal{L}\, S^{-1/2}$,
i.e.\ $P_{1,0}$. ULoRA makes this choice explicit and continuous;
we initialize the adapters from the top-$r$ subspace of
$P_{\alpha,\beta}$, computed efficiently in a projected Rayleigh--Ritz basis as described next.
 
\subsection{Curvature Estimation and Gradient Projection}
For a linear layer with pretrained weight $W_0 \in \mathbb{R}^{d_\text{out} \times d_\text{in}}$, let $h_i \in \mathbb{R}^{d_\text{in}}$ be the layer input and $\delta_i \in \mathbb{R}^{d_\text{out}}$ be the pre-activation output gradient for sample $i$. On a small initialization batch, the K-FAC approximation \cite{martens2015optimizing} to the Hessian of the loss yields two Kronecker factors:
\begin{equation}
S = \frac{1}{n}\sum_i h_i h_i^\top \in \mathbb{R}^{d_\text{in} \times d_\text{in}},
\qquad
T = \sum_i \delta_i \delta_i^\top \in \mathbb{R}^{d_\text{out} \times d_\text{out}},
\end{equation}
where $S$ captures the covariance of layer inputs and $T$ captures the covariance of output gradients across the batch. We also collect the per-feature second moments:
\begin{equation}
d_S[j] = \frac{1}{n}\sum_i h_{ij}^2,
\qquad
d_T[k] = \frac{1}{n}\sum_i \delta_{ik}^2,
\end{equation}
which are the diagonal entries of $S$ and $T$. Note that $d_S$ and $d_T$ are the per-coordinate second moments used by Adam as a diagonal curvature proxy. Materializing $P_{\alpha,\beta}(G)$ at full size would be as expensive as a full fine-tuning step, so all computation is carried out in a low-dimensional projected space. We compute a rank-$s$ ($s = r + \text{oversample}$) Rayleigh--Ritz approximation of $S$ and $T$, obtaining the principal eigenvectors and eigenvalues:
\begin{equation}
S \approx U_S D_S U_S^\top,
\qquad
T \approx U_T D_T U_T^\top.
\end{equation}
The projected gradient is defined as:
\begin{equation}
\hat{F} = U_T^\top \, \nabla_W \mathcal{L} \, U_S \in \mathbb{R}^{s \times s}.
\end{equation}
This matrix represents the loss gradient compressed into the Kronecker subspace: it indicates which combinations of input and output directions contain the strongest gradient signal. The two preconditioning exponents of Eq.~\ref{eq:family} are then applied in this projected space.
 
\subsection{The $\beta$ Parameter: Adam-like Diagonal Preconditioning}
 
Before forming $\hat{F}$, we rescale the basis vectors $U_S$ and $U_T$ per feature using the diagonal statistics $d_S$ and $d_T$:
\begin{equation}
\tilde{U}_S = (d_S + \varepsilon_S)^{-\beta/2} \odot U_S,
\qquad
\tilde{U}_T = (d_T + \varepsilon_T)^{-\beta/2} \odot U_T,
\end{equation}
where $\odot$ denotes elementwise row scaling and $\varepsilon > 0$ is a small numerical stabilizer. Features with large empirical variance (large $d_S[j]$ or $d_T[k]$) are downweighted proportionally. This directly mirrors the adaptive step size in Adam: just as Adam divides the gradient by the root mean square of past gradients per coordinate, ULoRA divides the basis vectors by the root mean square of layer inputs and output gradients per feature. The projected gradient using these scaled bases is:
\begin{equation}
\hat{F}^{\beta} = \tilde{U}_T^\top \, \nabla_W \mathcal{L} \, \tilde{U}_S.
\end{equation}
When $\beta = 0$, no diagonal correction is applied and $\hat{F}^0 = \hat{F}$. When $\beta = 1$, each feature dimension is normalized by its empirical standard deviation.
 
\subsection{The $\alpha$ Parameter: Spectral Whitening Power}
 
Given the K-FAC eigenvalues $D_S$ and $D_T$, we apply spectral scaling to the projected gradient:
\begin{equation}
F^{\alpha,\beta} = D_T^{-\alpha/2} \, \hat{F}^{\beta} \, D_S^{-\alpha/2},
\end{equation}
where each row $k$ is scaled by $\lambda_{T,k}^{-\alpha/2}$ and each column $j$ by $\lambda_{S,j}^{-\alpha/2}$. This reweights gradient directions by the inverse power of their curvature: directions residing in sharp regions of the loss landscape (large eigenvalues) are shrunk, while directions in flat regions (small eigenvalues, but broad functional influence) are amplified. The effect of $\alpha$ is as follows:
\begin{itemize}
    \item $\alpha = 0$: No spectral whitening. Only the $\beta$ diagonal correction is applied. Gradient directions are weighted by their raw magnitude in the K-FAC subspace.
    \item $\alpha = 0.5$: Quarter-power scaling per side, matching the preconditioning strength of Shampoo \cite{gupta2018shampoo}.
    \item $\alpha = 1$, $\beta = 0$: Inverse-square-root whitening. Recovers CG-LoRA \cite{zheng2026cg-LoRA} exactly.
    \item $\alpha = 2$: Full inverse-curvature scaling, corresponding to the K-FAC natural gradient; increasingly penalizes high-curvature directions and amplifies estimation noise in flat ones.
\end{itemize}

\subsection{Cross-Entropy Correction}
The whitening above treats output space as Euclidean, which is exact
for squared loss, whose Hessian with respect to the network output is
the identity. Cross-entropy instead induces the output-space curvature
$\Lambda = \mathrm{diag}(p) - pp^\top$, where $p$ is the softmax
output \cite{martens2015optimizing}, so distances in output space must
be measured in this metric. We
accumulate this curvature at the layer output over the initialization
batch, $\hat{H}_\text{out}$, project it into the whitened basis,
$\Phi = D_T^{-\alpha/2}\, \tilde{U}_T^\top \hat{H}_\text{out}
\tilde{U}_T\, D_T^{-\alpha/2}$, and whiten once more:
$G = \Phi^{-1/2} F^{\alpha,\beta}$. For squared loss,
$G = F^{\alpha,\beta}$ directly; for details see \cite{zheng2026cg-LoRA}.
 
\subsection{Symmetric Low-Rank Initialization} 
 
Given the final matrix $G$, we compute its rank-$r$ SVD and back-project the singular vectors to the original parameter space:
\begin{equation}
L_r = \tilde{U}_T \, D_T^{-\alpha/2} \, U_r,
\qquad
R_r = \tilde{U}_S \, D_S^{-\alpha/2} \, V_r,
\end{equation}
where $U_r$ and $V_r$ are the top-$r$ left and right singular vectors of $G$. We then apply a QR and tiny-SVD balanced factorization to distribute singular values equally between $A_0$ and $B_0$, ensuring neither adapter dominates. Concretely, let $Q_L R_L = \text{QR}(L_r)$ and $Q_R R_R = \text{QR}(R_r)$. We form the $r \times r$ core matrix $M = R_L \, \text{diag}(D_r) \, R_R^\top$ and compute its SVD $M = U_M S_M V_M^\top$. Then:
\begin{equation}
B_0 = Q_L U_M \sqrt{S_M},
\qquad
A_0 = \left( Q_R V_M \sqrt{S_M} \right)^\top.
\end{equation}
Finally, we normalize both adapters: $A_0 \leftarrow \frac{d_\text{out}^{1/4}}{\gamma} A_0$ and $B_0 \leftarrow \frac{d_\text{out}^{1/4}}{\gamma} B_0$, with $\gamma = 16$. The entire procedure never forms a matrix of size $d_\text{out} \times d_\text{in}$; all computation stays in the projected space of dimension $s = r + \text{oversample}$.
 

\subsection{Initialization Workflow} 
The full step-by-step pipeline of ULoRA initialization is summarized in Table~\ref{tab:workflow}. After initialization, standard LoRA fine-tuning proceeds:
\begin{equation}
W = W_0 + \tfrac{1}{\sqrt{r}} (B_0 + \Delta B)(A_0 + \Delta A),
\end{equation}
updating only $\Delta A$ and $\Delta B$.

\subsection{LoRA-Auto}
One way to set $(\alpha, \beta)$ is grid search, but can the operating
point be chosen from data, without any search? Our analysis suggests it
can: the risk of strong whitening is amplifying noise-dominated flat
directions, and the usefulness of diagonal scaling depends on the
covariance actually being near-diagonal. Both properties
are measurable per layer from quantities already computed during
initialization.
For each side $\Sigma \in \{S, T\}$ with eigenvalues $\{d_i\}$, we
summarize the spectrum by its effective rank
$r_\text{eff} = (\sum_i d_i)^2 / \sum_i d_i^2$ and condition number
$\kappa = \max_i(d_i)/(\min_i(d_i) + \epsilon)$, and set
\begin{equation}
\alpha^{*} = \frac{2}{\,1 + \ln(\kappa) / \big(\ln(s)\, r_\text{eff}\big)\,},
\end{equation}
so that well-conditioned, high-effective-rank spectra receive strong
whitening ($\alpha^{*} \!\to\! 2$) while spiky, ill-conditioned spectra,
where inverse-power scaling would amplify estimation noise, are
whitened conservatively. For $\beta$, we compute the off-diagonal energy
ratio of the projected covariance $B \in \{B_\text{in}, B_\text{out}\}$,
$\omega = \big(\|B\|_F^2 - \|\mathrm{diag}(B)\|_F^2\big) / \|B\|_F^2$,
and set $\beta^{*} = (1-\omega)^{2}$; diagonal
preconditioning is applied only to the extent that the covariance is
empirically diagonal. Each adapted layer thus receives its own
$(\alpha^{*}_S, \alpha^{*}_T, \beta^{*}_S, \beta^{*}_T)$ at zero search
cost, navigating a per-layer design space no single global grid point
can reach.

\begin{table*}[t]
\centering
\small
\begin{tabular}{clp{12cm}}
\toprule
Step & Phase & Operation \\
\midrule
1 & Curvature Estimation & Run forward--backward pass on the initialization batch; accumulate K-FAC factors $S, T$ and per-feature diagonal statistics $d_S, d_T$. \\
2 & Rayleigh--Ritz & Compute top-$s$ eigendecomposition: $S \approx U_S D_S U_S^\top$ and $T \approx U_T D_T U_T^\top$ ($s = r + \text{oversample}$). \\
3 & ULoRA-Auto (if enabled) & Per layer and side: from the eigenvalues $\{d_i\}$ of $D_S, D_T$ compute $r_\text{eff} = (\sum_i d_i)^2 / \sum_i d_i^2$ and $\kappa = \max_i(d_i)/(\min_i(d_i)+\epsilon)$; set $\alpha^{*} = 2 / \big(1 + \ln(\kappa)/(\ln(s)\, r_\text{eff})\big)$. From the projected covariances $B_\text{in}, B_\text{out}$ compute the off-diagonal energy ratio $\omega = (\|B\|_F^2 - \|\mathrm{diag}(B)\|_F^2)/\|B\|_F^2$; set $\beta^{*} = (1-\omega)^{2}$. In manual mode this step is skipped and the given global $(\alpha, \beta)$ is used. \\
4 & Diagonal Scaling ($\beta$) & Construct $\beta$-preconditioned bases: $\tilde{U}_S = (d_S + \varepsilon_S)^{-\beta_S/2} \odot U_S$ and $\tilde{U}_T = (d_T + \varepsilon_T)^{-\beta_T/2} \odot U_T$. \\
5 & Gradient Projection & Accumulate preconditioned projected gradient: $\hat{F}^{\beta} = \tilde{U}_T^\top \, \nabla_W \mathcal{L} \, \tilde{U}_S$. \\
6 & Spectral Whitening ($\alpha$) & Apply spectral scaling: $F^{\alpha,\beta} = D_T^{-\alpha_T/2} \hat{F}^{\beta} D_S^{-\alpha_S/2}$. \\
7 & Fisher Correction & If cross-entropy loss, compute $\Phi = D_T^{-\alpha_T/2} \tilde{U}_T^\top \hat{H}_\text{out} \tilde{U}_T D_T^{-\alpha_T/2}$ and set $G = \Phi^{-1/2} F^{\alpha,\beta}$; else $G = F^{\alpha,\beta}$. \\
8 & Low-Rank SVD & Compute rank-$r$ SVD of $G$: $G \approx U_r \,\mathrm{diag}(D_r)\, V_r^\top$, with singular values $D_r$. \\
9 & Back-projection & Map subspace modes to model space: $L_r = \tilde{U}_T D_T^{-\alpha_T/2} U_r$ and $R_r = \tilde{U}_S D_S^{-\alpha_S/2} V_r$. \\
10 & Balanced Realization & Decompose via QR: $Q_L R_L = \text{QR}(L_r)$, $Q_R R_R = \text{QR}(R_r)$. SVD of core $R_L \, \mathrm{diag}(D_r) \, R_R^\top = U_M S_M V_M^\top$. Set $B_0 = Q_L U_M \sqrt{S_M}$, $A_0 = (Q_R V_M \sqrt{S_M})^\top$. \\
11 & Global Normalization & Rescale adapters: $A_0 \leftarrow \frac{d_\text{out}^{1/4}}{\gamma} A_0$, $B_0 \leftarrow \frac{d_\text{out}^{1/4}}{\gamma} B_0$. \\
\bottomrule
\end{tabular}
\caption{Step-by-step workflow of ULoRA initialization per layer. Step 3 is enabled only in ULoRA-Auto, which selects $(\alpha^{*}_S, \alpha^{*}_T, \beta^{*}_S, \beta^{*}_T)$ per layer and per side from the spectral shape ($r_\text{eff}, \kappa$) and off-diagonal energy ($\omega$) of the estimated covariances; steps 4--9 then use these per-side exponents.}
\label{tab:workflow}
\end{table*}

\section{Experiments}
\paragraph{Models and tasks.}
We evaluate ULoRA in two settings. \textbf{NLU:} we fine-tune RoBERTa-base \cite{liu2019RoBERTa} and T5-base \cite{raffel2020exploring} on five GLUE tasks \cite{wang2018glue} (CoLA, MNLI, QNLI, SST-2, MRPC); for T5 we score each class by teacher-forcing its label token. \textbf{NLG:} we fine-tune LLaMA 2-7B \cite{touvron2023llama2} on three tasks: mathematical reasoning, trained on 100K MetaMathQA samples \cite{Yu2023} and evaluated on GSM8K \cite{cobbe2021training} with regex-extracted answer accuracy; code generation, trained on 100K Code-Feedback samples \cite{Wei2023} with explanations removed and evaluated on HumanEval \cite{Chen2021} with Pass@1; and question answering, trained on a filtered 52K WizardLM subset \cite{Xu2023} and evaluated on MMLU \cite{Hendrycks2020} with answer accuracy.
 
\paragraph{Baselines.}
We compare against rsLoRA \cite{kalajdzievski2023rank}, LoRA+ \cite{hayou2024LoRA+}, PiSSA \cite{meng2024pissa}, LoRA-GA \cite{wang2024LoRAga}, LoRA-One \cite{zhang2025LoRA1}, and CG-LoRA \cite{zheng2026cg-LoRA}, as well as full fine-tuning.
 
\paragraph{Configurations.}
We set rank $r = 8$, no dropout, and $\gamma = 16$. For RoBERTa and T5, we adapt query, key, value, and dense projection layers. For LLaMA, we adapt all linear layers within the transformer backbone, excluding the final language modeling head.
 
\paragraph{Training.}
All models are trained for one epoch with AdamW ($\beta_1 = 0.9$, $\beta_2 = 0.999$, weight decay $0$), a cosine learning rate schedule with 3\% warmup, batch size 32. We employ a mixed-precision framework where the LLaMA backbone is maintained in BF16, and the LoRA's $A$ and $B$ matrices are cast to FP32 across all evaluated architectures. Each configuration is run with 3 random seeds, and we report the average performance.
 
\paragraph{Learning rate sweep.}
For RoBERTa and T5, we search the learning rate over $\{2 \times 10^{-5}, 5 \times 10^{-5}, 10^{-4}, 2 \times 10^{-4}, 5 \times 10^{-4}, 8 \times 10^{-4}, 10^{-3}\}$ and report the best. For LLaMA, we tune the learning rate with Optuna \cite{agrawal2020optuna} since learning rate strongly affects LoRA performance \cite{lee2026learning}.

\paragraph{ULoRA grid.}
We sweep $\alpha \in \{0.0, 0.5, 1.0, 1.5, 2.0\}$ and $\beta \in \{0.0, 0.5, 1.0\}$, giving 15 configurations per task.
 
\paragraph{Compute infrastructure.}
Experiments for RoBERTa-base and T5-base were conducted on NVIDIA RTX 6000 Pro Blackwell GPUs while for LLaMA 2-7B, we used NVIDIA H200 GPUs.
 
\subsection{Results}
 
Two findings emerge consistently across all experiments. First, preconditioning strength matters: taken as an oracle upper bound of the family, the best point in the $(\alpha, \beta)$ family matches or exceeds baselines on 11 of 13 comparisons and matches or exceeds full fine-tuning on all five RoBERTa GLUE tasks. Second, and more surprising, the right strength is not what existing methods assume: the optimum is task- and model-dependent, frequently lies away from both published endpoints, no preconditioning in LoRA-GA and LoRA-One and full inverse-square-root whitening in CG-LoRA, and neither endpoint is optimal in the majority of settings. The claim we defend is therefore not that one configuration wins everywhere, which our own data rules out, but that the preconditioning exponent is a design variable with measurable effects.

\begin{table}[t]
\centering
\footnotesize
\setlength{\tabcolsep}{3pt}
\begin{tabular}{lccccc}
\toprule
\textbf{Method} & \textbf{CoLA} & \textbf{MNLI} & \textbf{QNLI} & \textbf{SST-2} & \textbf{MRPC} \\
\midrule
FF       & 81.0{\tiny$\pm$0.6} & 86.8{\tiny$\pm$0.2} & 92.2{\tiny$\pm$0.2} & 93.8{\tiny$\pm$0.0} & 86.8{\tiny$\pm$0.5} \\
\midrule
LoRA     & 80.2{\tiny$\pm$0.4} & 86.4{\tiny$\pm$0.1} & \underline{92.0}{\tiny$\pm$0.1} & \underline{93.8}{\tiny$\pm$0.1} & 83.5{\tiny$\pm$1.6} \\
LoRA+    & 79.0{\tiny$\pm$0.7} & 86.2{\tiny$\pm$0.2} & 91.3{\tiny$\pm$0.0} & 93.7{\tiny$\pm$0.2} & 85.5{\tiny$\pm$1.5} \\
rsLoRA   & \underline{80.7}{\tiny$\pm$0.5} & 86.4{\tiny$\pm$0.1} & 91.9{\tiny$\pm$0.1} & 93.2{\tiny$\pm$0.4} & 85.3{\tiny$\pm$1.6} \\
PISSA    & 80.6{\tiny$\pm$0.6} & \underline{86.5}{\tiny$\pm$0.1} & 91.9{\tiny$\pm$0.1} & 93.4{\tiny$\pm$0.2} & 85.5{\tiny$\pm$0.6} \\
LoRA-GA  & 77.5{\tiny$\pm$0.3} & 85.2{\tiny$\pm$0.2} & 90.6{\tiny$\pm$0.3} & 93.3{\tiny$\pm$0.1} & 84.9{\tiny$\pm$1.4} \\
LoRA-One  & 80.1{\tiny$\pm$0.7} & 86.4{\tiny$\pm$0.2} & 91.4{\tiny$\pm$0.2} & \underline{93.8}{\tiny$\pm$0.2} & 84.7{\tiny$\pm$0.9} \\
CG-LoRA  & \underline{80.7}{\tiny$\pm$0.6} & \underline{86.5}{\tiny$\pm$0.1} & \textbf{92.1}{\tiny$\pm$0.1} & \textbf{93.9}{\tiny$\pm$0.3} & \textbf{87.0}{\tiny$\pm$0.9} \\
\midrule
ULoRA (UB)$^{*}$ & 81.1{\tiny$\pm$0.4} & 86.8{\tiny$\pm$0.1} & 92.2{\tiny$\pm$0.2} & 94.5{\tiny$\pm$0.5} & 87.0{\tiny$\pm$0.8} \\
{\scriptsize$(\alpha,\beta)$} & {\scriptsize(1.0,1.0)} & {\scriptsize(1.0,0.0)} & {\scriptsize(1.5,0.5)} & {\scriptsize(1.5,0.5)} & {\scriptsize(1.0,0.0)} \\
 
ULoRA-Auto & \textbf{80.8}{\tiny$\pm$0.5} & \textbf{86.6}{\tiny$\pm$0.1} & \textbf{92.1}{\tiny$\pm$0.2} & \textbf{93.9}{\tiny$\pm$0.6} & \underline{86.4}{\tiny$\pm$0.8} \\
\bottomrule
\end{tabular}
 
\caption{GLUE benchmark results for RoBERTa, averaged over 3 seeds (mean $\pm$ sd). Bold/underline = best/second best among deployable methods. $^{*}$ULoRA (UB) is an oracle upper bound; the best $(\alpha,\beta)$ was selected.}
\label{tab:glue-RoBERTa-results}
 
\end{table}

\begin{table}[t]
\centering
\footnotesize
\setlength{\tabcolsep}{3pt}
\begin{tabular}{lccccc}
\toprule
\textbf{Method} & \textbf{CoLA} & \textbf{MNLI} & \textbf{QNLI} & \textbf{SST-2} & \textbf{MRPC} \\
\midrule
FF       & 81.6{\tiny$\pm$0.8} & 86.2{\tiny$\pm$0.2} & 93.3{\tiny$\pm$0.0} & 94.4{\tiny$\pm$0.2} & 87.0{\tiny$\pm$0.7} \\
\midrule
LoRA     & 79.9{\tiny$\pm$2.1} & \underline{85.7}{\tiny$\pm$0.1} & \textbf{93.3}{\tiny$\pm$0.3} & 94.1{\tiny$\pm$0.3} & 83.5{\tiny$\pm$1.6} \\
LoRA+    & \underline{80.8}{\tiny$\pm$0.4} & 85.6{\tiny$\pm$0.0} & 93.1{\tiny$\pm$0.0} & \underline{94.3}{\tiny$\pm$0.4} & 85.7{\tiny$\pm$1.4} \\
rsLoRA   & 80.2{\tiny$\pm$0.3} & \underline{85.7}{\tiny$\pm$0.3} & \textbf{93.3}{\tiny$\pm$0.3} & 94.2{\tiny$\pm$0.3} & 85.1{\tiny$\pm$0.6} \\
PISSA    & 80.2{\tiny$\pm$0.7} & \textbf{85.8}{\tiny$\pm$0.0} & \underline{93.2}{\tiny$\pm$0.1} & \underline{94.3}{\tiny$\pm$0.1} & 85.3{\tiny$\pm$0.7} \\
LoRA-GA  & \underline{80.8}{\tiny$\pm$0.2} & \underline{85.7}{\tiny$\pm$0.1} & \textbf{93.3}{\tiny$\pm$0.1} & \textbf{94.4}{\tiny$\pm$0.1} & \textbf{86.3}{\tiny$\pm$0.5} \\
LoRA-One  & \textbf{80.9}{\tiny$\pm$0.1} & \textbf{85.8}{\tiny$\pm$0.1} & 93.1{\tiny$\pm$0.0} & 94.2{\tiny$\pm$0.0} & \underline{85.9}{\tiny$\pm$0.3} \\
CG-LoRA  & 80.6{\tiny$\pm$0.5} & \textbf{85.8}{\tiny$\pm$0.2} & \underline{93.2}{\tiny$\pm$0.2} & \textbf{94.4}{\tiny$\pm$0.2} & 85.3{\tiny$\pm$1.1} \\
\midrule
ULoRA (UB)$^{*}$ & 81.0{\tiny$\pm$0.4} & 85.8{\tiny$\pm$0.1} & 93.3{\tiny$\pm$0.2} & 94.5{\tiny$\pm$0.3} & 86.0{\tiny$\pm$0.9} \\
{\scriptsize$(\alpha,\beta)$} & {\scriptsize(0.5,0.5)} & {\scriptsize(0.0,0.0)} & {\scriptsize(1.5,0.0)} & {\scriptsize(0.0,0.0)} & {\scriptsize(0.0,1.0)} \\
 
ULoRA-Auto & 80.7{\tiny$\pm$0.2} & \textbf{85.8}{\tiny$\pm$0.1} & \textbf{93.3}{\tiny$\pm$0.1} & \underline{94.3}{\tiny$\pm$0.2} & 85.3{\tiny$\pm$0.3} \\
\bottomrule
\end{tabular}
 
\caption{GLUE benchmark results for T5, averaged over 3 seeds (mean $\pm$ sd). Bold/underline = best/second best among deployable methods. $^{*}$ULoRA (UB) is an oracle upper bound; the best $(\alpha,\beta)$ was selected.}
\label{tab:glue-t5-results}
\end{table}
 
\paragraph{RoBERTa on GLUE.}
Table~\ref{tab:glue-RoBERTa-results} shows that some point in the $(\alpha, \beta)$ family matches or exceeds every baseline, including full fine-tuning, on all five tasks: CoLA (81.1 vs.\ 81.0), MNLI (86.8, tie), QNLI (92.2, tie), SST-2 (94.5 vs.\ 93.8), and MRPC (87.0 vs.\ 86.8). More important than the margins is where these points lie: the selected exponents cluster at moderate whitening, $\alpha \in \{1.0, 1.5\}$, with $\beta > 0$ on three tasks, and on three of five tasks the optimum is not the CG-LoRA point $(1.0, 0.0)$. In this regime, encoder-only classification with a small initialization batch, the K-FAC estimates are comparatively clean, so amplifying flat but functionally influential directions pays off, and the diagonal $\beta$ correction absorbs heterogeneous feature scales.

\paragraph{T5 on GLUE.}
The T5 results (Table~\ref{tab:glue-t5-results}) are best interpreted as a saturation regime. With a tuned learning rate, vanilla LoRA already matches full fine-tuning on QNLI (93.3 vs.\ 93.3) and SST-2 (94.1 vs.\ 94.4), all methods fall within 0.2 points of each other on MNLI and QNLI, and the residual gaps on the small, high-variance tasks CoLA and MRPC are comparable to a single seed standard deviation (0.5 to 2.1). When tuned LoRA saturates to full fine-tuning accuracy, consistent with \cite{lee2026learning}, initialization has little leverage, and no method can meaningfully separate from the pack; the family's oracle upper bound is accordingly at or near the top, with margins we do not consider significant. The informative signal is instead where the optimum lands: it collapses toward weak preconditioning, with $(\alpha, \beta) = (0.0, 0.0)$ selected on MNLI and SST-2, possibly because gradients through cross-attention and a large output softmax yield noisier K-FAC factors. This is consistent with the framework rather than against it; where there is no headroom and curvature estimates are unreliable, the family correctly selects little to no preconditioning .

 
\begin{table}[t]
\centering
\footnotesize
\setlength{\tabcolsep}{3pt}
\begin{tabular}{lccc}
\toprule
\textbf{Method} & \textbf{GSM8K} & \textbf{HumanEval} & \textbf{MMLU} \\
\midrule
FF        & 54.54{\tiny$\pm$0.84} & 23.78{\tiny$\pm$0.86} & 45.98{\tiny$\pm$0.21} \\
\midrule
LoRA      & \underline{57.29}{\tiny$\pm$1.42} & 26.22{\tiny$\pm$1.00} & 44.35{\tiny$\pm$0.76} \\
LoRA+     & 50.06{\tiny$\pm$0.74} & 26.02{\tiny$\pm$1.75} & \underline{45.35}{\tiny$\pm$0.28} \\
rsLoRA    & 57.06{\tiny$\pm$0.29} & 26.42{\tiny$\pm$2.46} & 43.44{\tiny$\pm$0.28} \\
PISSA     & 54.79{\tiny$\pm$0.44} & \underline{26.63}{\tiny$\pm$1.25} & 40.84{\tiny$\pm$0.59} \\
LoRA-GA   & 50.72{\tiny$\pm$0.33} & 24.80{\tiny$\pm$1.04} & 44.36{\tiny$\pm$0.63} \\
LoRA-One   & 55.85{\tiny$\pm$0.13} & 23.78{\tiny$\pm$0.86} & 45.01{\tiny$\pm$0.37} \\
CG-LoRA   & \textbf{57.70}{\tiny$\pm$0.84} & 26.02{\tiny$\pm$0.76} & 44.97{\tiny$\pm$0.60} \\
\midrule
ULoRA (UB)$^{*}$ & 57.97{\tiny$\pm$0.79} & 28.46{\tiny$\pm$1.6} & 45.27{\tiny$\pm$0.50} \\
{\scriptsize$(\alpha,\beta)$} & {\scriptsize(0.0,0.0)} & {\scriptsize(2.0,0.5)} & {\scriptsize(1.5,1.0)} \\
 
ULoRA-Auto & 57.16{\tiny$\pm$0.11} & \textbf{27.24}{\tiny$\pm$0.29} & \textbf{45.43}{\tiny$\pm$0.18} \\
\bottomrule
\end{tabular}
 
\caption{Benchmark results for LLaMA-2-7B, averaged over 3 seeds (mean $\pm$ sd). Bold/underline = best/second best among deployable methods. $^{*}$ULoRA (UB) is an oracle upper bound; the best $(\alpha,\beta)$ was selected.}
\label{tab:llama2-7b-results}
\end{table}
 
\paragraph{LLaMA 2-7B.}
The generation benchmarks (Table~\ref{tab:llama2-7b-results}) push this further. The best configuration is (0.0,0.0) on GSM8K (57.97), (2.0,0.5) on HumanEval (28.46), and (1.5,1.0) on MMLU, so the optimum varies across tasks even within one backbone. At the 7B scale, single-batch curvature estimates over large hidden dimensions are the noisiest of our settings, and for GSM8K, the raw gradient subspace is the safest choice. Note that ULoRA at $(0.0, 0.0)$ still outperforms LoRA-GA (50.72 on GSM8K) and LoRA-One (55.85), because the rest of the pipeline differs: the Rayleigh--Ritz projection, the cross-entropy Fisher correction, and the balanced realization apply regardless of the exponents. Consistent with prior reports, several LoRA variants exceed full fine-tuning on GSM8K and HumanEval; the low-rank constraint acts as a regularizer in single-epoch instruction tuning.

\begin{figure}[h]
    \centering
    \includegraphics[width=1.0\linewidth]{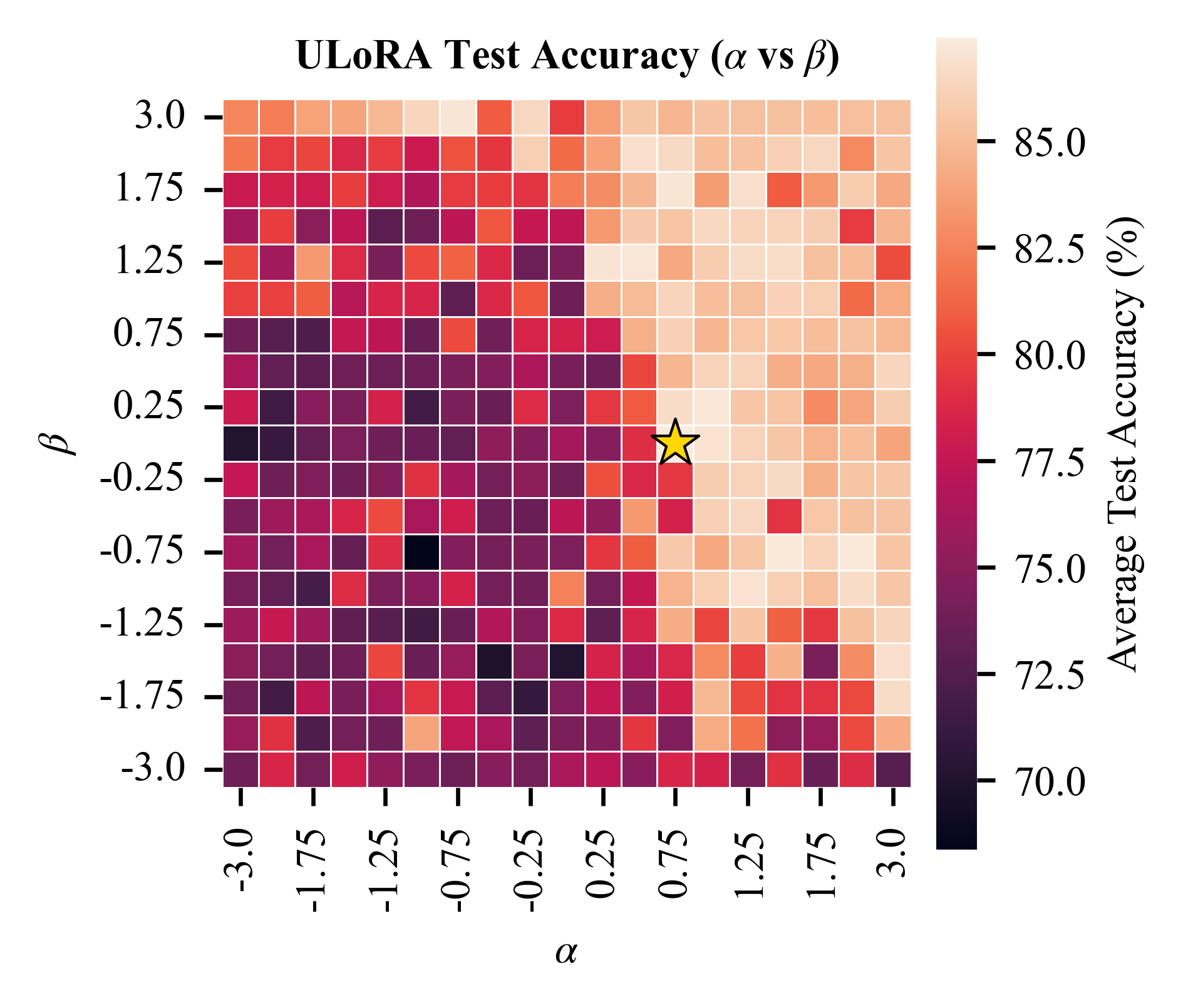}
    \caption{Heatmap of the fine grid search for $\alpha, \beta \in [-3, +3]$ on GLUE MRPC, averaged across 3 seed runs. The maximum accuracy is marked by a star.}
    \label{fig:cnt}
\end{figure}
 
\paragraph{Accuracy landscape.}
Figure~\ref{fig:cnt} maps test accuracy on MRPC over the extended range $\alpha, \beta \in [-3, +3]$. The landscape is highly asymmetric in $\alpha$: negative values, which amplify high-curvature directions instead of suppressing them, degrade accuracy by up to 15 points, while $\alpha>0.5$ forms a broad plateau of high accuracy within which $\beta$ has a mild effect. The maximum lies in the interior near $(0.75, -0.25)$, at none of the published endpoints. Two practical points follow: the exponent is forgiving once past the transition, so precise tuning is rarely necessary, but the existence of a sharp failure region shows that the choice is not free either.

\paragraph{ULoRA-Auto.}
Grid search multiplies initialization cost, so we also evaluate ULoRA-Auto, which sets $(\alpha, \beta)$ per layer from spectral statistics the pipeline already computes: the effective rank and condition number of the Ritz eigenvalue spectra determine a decoupled whitening exponent for the input and output sides, and the off-diagonal energy ratio of the projected covariances sets an exponentially damped $\beta$. This rule backs off whitening exactly when the curvature estimate looks ill-conditioned or structurally noisy. It is a heuristic rather than a derived optimum, but it performs remarkably well for one: with zero tuning it ranks at or near the top among deployable methods on every RoBERTa task, stays within 0.1 to 0.6 points of the oracle grid search throughout, and is the single best method on MMLU (45.43), where it beats its own grid-searched parent. In practice, ULoRA-Auto delivers most of the benefit of the full family at the cost of one initialization pass, and it provides direct evidence that the right exponent is predictable from spectral statistics available at initialization, not only findable by search.
 
\begin{figure}[h]
    \centering
    \includegraphics[width=1.0\linewidth]{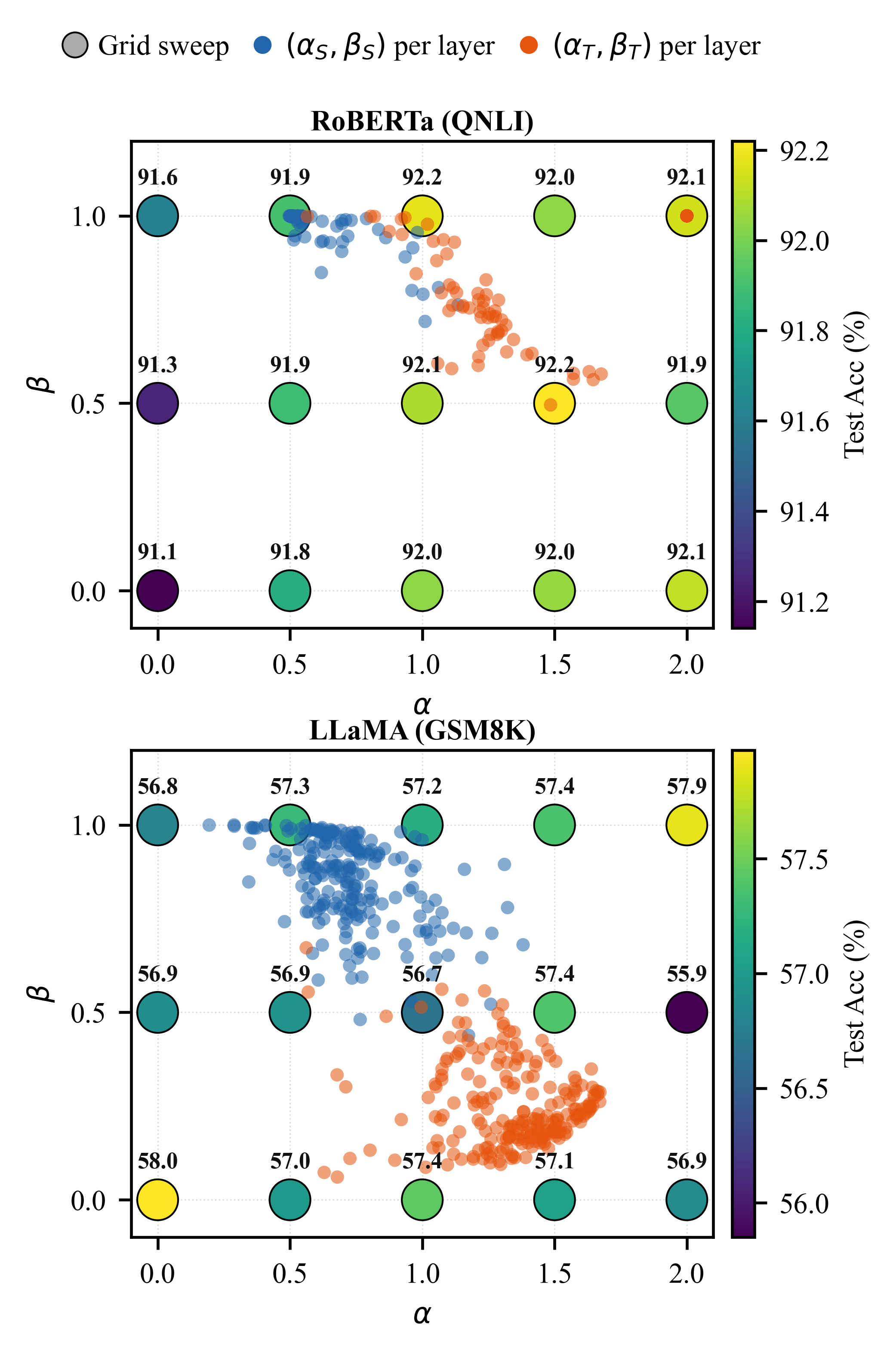}
    \caption{Per-layer exponents selected by ULoRA-Auto (dots) overlaid on the global grid sweep (circles, mean test accuracy) for QNLI (top) and GSM8K (bottom).}
    \label{fig:auto}
\end{figure}

Auto's per-layer selections separate cleanly by side ($\beta_S$ near 1, $\beta_T$ low), matching the measured off-diagonal energy of each (Figure~\ref{fig:auto}). On QNLI, the selections land in the
grid's best region and recover near-optimal accuracy at zero search cost. On GSM8K, where the grid favors the raw-gradient corner, Auto prescribes moderate whitening instead: the diagnostics measure estimation quality, not task-level utility.
 
\paragraph{Limitations.}
The grid-searched ULoRA numbers select $(\alpha, \beta)$ per task on test metrics, an oracle protocol; ULoRA-Auto is the deployable variant and should be read as our single-configuration result. The grid-searched family further assumes a single global $(\alpha, \beta)$ shared across all layers and both Kronecker sides, adopted only for search tractability, since a per-layer grid search would multiply the already substantial sweep cost by the number of adapted layers; ULoRA-Auto lifts this assumption per layer and side at no search cost, which is why it can exceed the best global-grid point (e.g., on MMLU). The ULoRA-Auto rule is empirically motivated, and we provide no theoretical characterization of the optimal exponent as a function of curvature-estimation noise. Experiments cover rank 8, single-epoch training, and models up to 7B parameters on NLP benchmarks; behavior at higher ranks, longer training, and other modalities is untested. Initialization adds one forward-backward pass and per-layer eigendecompositions, a small but nonzero overhead over vanilla LoRA.
 
\section{Conclusion}
We showed that gradient-based LoRA initialization methods, which have so far been developed as separate techniques, are points on a single two-parameter continuum of preconditioned gradients, and that this reframing has empirical teeth. Under a full learning-rate sweep across two encoders and a 7B decoder, and treated as an oracle upper bound, tuned points in the family match or exceed every LoRA baseline on 11 of 13 comparisons and match or exceed full fine-tuning on all five RoBERTa GLUE tasks; the deployable ULoRA-Auto recovers most of this gain with no tuning. The deeper result is where those points lie: the optimal preconditioning strength is systematically task- and model-dependent, tracks the reliability of the curvature estimate, and coincides with a published endpoint only occasionally. Fixed choices, whether no whitening or full whitening, leave accuracy on the table in most settings, and the extended landscape shows a genuine failure region that a fixed choice cannot guard against. ULoRA-Auto demonstrates that the right strength can be read off spectral statistics at initialization, recovering most of the gain of an oracle grid search with no tuning. We view the specific family studied here as a first instantiation rather than the final word; making the preconditioning decision explicit, and continuous.

\bibliography{references}


\end{document}